\documentclass[letterpaper, 10 pt, conference]{ieeeconf}  

\IEEEoverridecommandlockouts                              

\usepackage{graphics} 
\usepackage{epsfig} 
\usepackage{mathptmx} 
\usepackage{times} 
\usepackage{amsmath} 
\usepackage{amssymb}  

\usepackage{xcolor}
\usepackage{hyperref}
\usepackage{caption}
\usepackage{subcaption}
\usepackage{booktabs}
\usepackage{cite}
\let\labelindent\relax
\usepackage{enumitem}
\usepackage{float}
\setlist[enumerate]{itemsep=2pt, topsep=2pt, parsep=0pt, partopsep=0pt}
\title{\LARGE\bf
Differentiable Dynamics for Autonomous Micro-Mobility Navigation
}

\author{Grace Cai$^{1}$, Joey Lee$^{1}$, Nithin Parepally$^{1}$, Laura Zheng$^{1}$, Ming C. Lin$^{1}$
\thanks{$^{1}$The authors are with the Department of Computer
Science, University of Maryland at College Park, MD,
U.S.A.
        Email: glcai@umd.edu, \{jlee27, nparepa\}@terpmail.umd.edu, \{lyzheng, lin\}@umd.edu} %
}

\begin{document}

\maketitle
\thispagestyle{empty}
\pagestyle{empty}

\begin{abstract}
Autonomous micro-mobility vehicles (MMVs) such as wheelchairs, scooters, and bicycles have the potential to improve mobility access and support safe low-speed transportation in pedestrian-shared spaces. Achieving MMV autonomy will require realistic, predictable MMV motion. However, many existing autonomous vehicle stacks rely on simplified kinematic models that fail to capture key MMV characteristics such as tire slip, friction, and wheel layouts, limiting realism and gradient-based optimization. In this paper, we explore differentiable formulations of dynamics models for autonomous micro-mobility systems. We first construct DiffKBM, a differentiable version of the classical kinematic bicycle model (KBM). Then, we introduce DiffGM3, a differentiable formulation of the General Micro-Mobility Model (GM3), a unified tire-based dynamics formulation for micro-mobility vehicles that supports a wide range of MMV configurations. DiffKBM and DiffGM3 enable end-to-end differentiable optimization through MMV dynamics, making them suitable for integration into differentiable autonomy stacks. We evaluate these dynamics models in both open-loop and closed-loop settings: (1) open-loop trajectory matching, where DiffKBM and DiffGM3 are integrated as a dynamics layer within DiffStack~\cite{karkus2023diffstack} and optimized to reproduce real-world MMV trajectories, and (2) closed-loop autonomous navigation, where DiffKBM and DiffGM3 are paired with a differentiable model predictive control (MPC) controller in CrowdNav~\cite{chen2019crowd} pedestrian scenarios. In the open-loop setting, DiffGM3 outperforms DiffKBM in reproducing trajectories with improvements in ADE and NLL across bicycle, scooter, and motorcycle modes, and reductions in planning loss for bicycle and motorcycle trajectories. We also find that, in closed-loop settings, DiffGM3 improves on DiffKBM's CrowdNav performance by producing 55\% fewer collisions and a 75\% lower discomfort frequency for the bicycle mode.

\end{abstract}

\section{Introduction}
Autonomous micro-mobility vehicles (MMVs) such as wheelchairs, electric scooters, and bicycles are emerging as a promising mobility option for low-speed transportation in urban areas and environments with frequent mixed-traffic interactions, including campuses and plazas \cite{CORETTISANCHEZ2024101236, wu2025towards}. Beyond full autonomy, partial autonomy capabilities such as collision avoidance, assisted steering, and speed control can provide meaningful safety benefits \cite{fraudet2024swadapt2, yuan2023safety,ceravolo2017model}. Many applications also require navigation in pedestrian-shared spaces which motivates the need for predictable and physically realistic motion.

Pedestrian-shared environments present unique challenges for autonomous MMV navigation. MMVs must move through narrow corridors and dense crowds, respond to unpredictable human motion, and avoid other low-speed vehicles. Poorly represented dynamics can produce sudden changes in speed or heading, over-/under- steering, and uncomfortable stopping distances, increasing the risk of unsafe interactions \cite{baudry2018taking}. 

Many existing autonomous navigation stacks, however, rely on simplified vehicle models such as the unicycle or kinematic bicycle model (KBM) \cite{karkus2023diffstack, polack2017kinematic, liu2024height, ghosn2024hybrid}. These models fail to consider important MMV-specific characteristics, including tire slip, friction, rider lean, and wheel layout. Such simplifications not only limit physical realism but also the expressiveness of MMV behavior needed for predictable motion.

Differentiable autonomy stacks enable gradient-based optimization through perception, planning, control, and dynamics, allowing model parameters and cost weights to be learned directly from data \cite{NEURIPS2024_c592fc7e, oshin2023differentiable, lidec2024end}. DiffStack \cite{karkus2023diffstack} exemplifies this approach but relies on a dynamically-extended unicycle model \cite{lavalle2006planning}, which cannot capture complex MMV dynamics.

The General Micro-mobility Model (GM3) \cite{cai2025gm3} addresses this gap by introducing a unified dynamics formulation based on the {\em generalized} tire-brush model that supports a wide range of MMV configurations, including bicycles, scooters, skateboards, tricycles, and carts \cite{cai2025gm3}. It models longitudinal and lateral tire forces, aligning moments, load transfer, camber, and lean, which allow for physically realistic motion across diverse platforms. However, GM3 was originally formulated as a non-differentiable model, preventing its direct use in gradient-based optimization pipelines.

These prior works motivate a differentiable, physically grounded dynamics model for micro-mobility vehicles. In this effort, we reconstruct two dynamics models, the classical Kinematic Bicycle Model (KBM) and the General Micro-mobility Model (GM3), into differentiable formulations, DiffKBM and DiffGM3, suitable for integration into differentiable autonomy stacks. 

DiffGM3 treats a subset of physically meaningful parameters as {\em learnable}, including coefficient of friction, half contact length, cornering stiffness, and moments of inertia.  We evaluate the effectiveness, correctness, and accuracy of DiffGM3 in two settings using DiffKBM as a baseline. First, in open-loop trajectory matching, we integrate DiffKBM and DiffGM3 as the dynamics component within DiffStack \cite{karkus2023diffstack} and reproduce ground-truth MMV trajectories from the Infrastructural Multi-Person Trajectory and Context (IMPTC) dataset \cite{hetzel2023imptc}. We find that DiffGM3 outperforms DiffKBM in reproducing trajectories in the open-loop setting because of its ability to realign itself with damping and gain coefficients.

Second, in closed-loop navigation, we pair DiffGM3 and DiffKBM with differentiable model predictive control (DiffMPC) and evaluate the resulting system in CrowdNav \cite{chen2019crowd} pedestrian scenarios comparing against a differentiable KBM baseline. Results show that DiffGM3 exceeds DiffKBM's performance on nearly all metrics. In particular, DiffGM3 displays more cautious behavior, leading to substantially fewer collisions and lower discomfort frequencies, while increasing navigation time only slightly relative to DiffKBM.

\section{Related Work}
Differentiable programming is being explored in the current state of the art in autonomous vehicle controllers, specifically with model predictive control (MPC), which utilizes vehicle dynamics to solve an optimization problem and choose the best control sequence at each step. Nachkov et al. train a policy in Waymax (which uses KBM) with analytic policy gradients and an MSE loss function ~\cite{nachkov2025autonomousvehiclecontrollers}. Karkus et al. perform MPC over a finite horizon using the dynamically-extended unicycle model~\cite{karkus2023diffstack}. While these methods are less computationally expensive than deep learning control methods, they rely on simple dynamics models like KBM that do not account for slip forces or extend to novel wheel layouts, limiting the expressivity of robot maneuvers. 

Other work utilizes tire models that simulate more detailed dynamics and integrates them with MPC. Carvalho et al. use the Fiala tire model with MPC for 4-wheeled vehicles with collision avoidance constraints~\cite{predCtrl2013}. The Fiala tire model is a simplified closed-form solution of the brush model that uses fewer physical parameters. The controller is able to keep the vehicle safe in scenarios with low friction and multiple obstacles. However, this work does not generalize to other wheel layouts.

Several works have specifically studied the dynamics and control for micro-mobility platforms such as bicycles, scooters, and powered wheelchairs. Persson et al. develop an MPC controller for a riderless bicycle that incorporates lean and steer balance dynamics for stable trajectory tracking~\cite{persson2021riderlessbicycle}. Asperti et al. model vertical dynamics of an electric kick scooter that accounts for the mechanical impedance of the driver~\cite{asperti2022kickscooter}. Baudry et al. model caster wheel behavior in powered wheelchair kinematics to show how neglecting caster wheel dynamics during direction changes can lead to unsafe situations such as collisions~\cite{baudry2018taking}. These works highlight the importance of modeling platform-specific dynamics and slip effects; however, they tend to rely on simplified models and are usually specialized to a single platform type.

\section{Preliminaries}
\label{sec:GM3}
The General Micro-mobility Model (GM3) is a generalized model based on the \textit{brush tire model}, which represents the tire tread as independent bristles, each of which deflects under slip and friction forces~\cite{PACEJKA201287}. Tables~\ref{tab:params} and~\ref{tab:input} include the minimum parameters and variables required to calculate slip forces for a single tire.

\begin{table}[h]
\vspace{5pt}
\caption{Physical Parameters in the tire brush model.}
\label{tab:params}
    \centering
    \begin{tabular}{clc}
    \toprule
    \textbf{Symbol} & \textbf{Description} & \textbf{Units} \\
\midrule
        $\ell$ & Half contact length & m \\
        $R$ & Tire radius & m \\
        $\gamma$ & Camber angle & rad \\
        $\mu$ & Coefficient of friction & -\\
        $c_p$ & Tread element stiffness per unit area & N/m$^2$ \\
    \bottomrule
    \end{tabular}
\end{table}

\begin{table}[h!]
    \caption{Input Variables in the tire brush model.}
    \label{tab:input}
    \vspace{-5pt}
    \centering
    \begin{tabular}{clc}
    \toprule
    \textbf{Symbol} & \textbf{Description} & \textbf{Units} \\
\midrule
        $F_z$ & Normal load & N \\
         $\Omega$ & Wheel angular velocity & rad/s \\
         $\delta$ & Steering angle & rad \\
         $r$ & Vehicle yaw rate & rad/s \\
         $\alpha$ & Slip angle & rad \\   
    \bottomrule
    \end{tabular}
\end{table}
The theoretical slip vector is defined as $\mathbf{\sigma}$ with practical slip quantities $\kappa=-\frac{V_{sx}}{V_x}$ and $\tan \alpha = -\frac{V_{sy}}{V_x}$ (where $V_x$ and $V_y$ are the linear longitudinal and lateral velocities, $V_{sx}$ and $V_{sy}$ are the linear longitudinal and lateral slip velocities, and $V_{r}=R\Omega$ is the rolling velocity).
\begin{equation}
    \mathbf{\sigma} 
    = -\frac{\mathbf{V_s}}{V_r}
    = \begin{pmatrix}
        \sigma_x \\
        \sigma_y
    \end{pmatrix} \\
    = \begin{pmatrix}
        \frac{\kappa}{1+\kappa} \\
        \frac{\tan \alpha}{1+\kappa}
    \end{pmatrix}
\end{equation}

When a tire follows a curved path, it experiences turn slip represented by $\varphi$. For a tire with camber angle $\gamma$, radius $R$, turning radius $r_{\text{turn}}$, and reduction factor $\varepsilon_{\gamma}$ (close to 0 for small tires and close to 1 for large tires), the turn slip is

\begin{equation}
\label{eq:turn_slip}
    \varphi = -\frac{1}{r_{\text{turn}}} + (1-\varepsilon_\gamma) \frac{\sin(\gamma)}{R}.
\end{equation}

The resulting {\em longitudinal force $F_x$, lateral force $F_y$, and aligning moment $M_z$} are as follows.
{\footnotesize
\begin{align}
    F_x &= 
    \begin{cases}
        \label{eq:Fx}
       \mu F_z \frac{\sigma_x}{\sigma} [3\psi\sigma-3(\psi\sigma)^2+(\psi \sigma)^3], & \sigma\le \frac{1}{\psi}\\
        \mu F_z \frac{\sigma_x}{\sigma} ,           & \sigma > \frac{1}{\psi}
    \end{cases} \\
    F_y &= 
    \begin{cases}
        \label{eq:Fy}
        \mu F_z [3\psi^*\sigma_{y}-3(\psi^*\sigma_{y})^2+(\psi^* \sigma_{y})^3] + \frac{2}{3}c_p\ell^3\varphi, & |\sigma_{y}|\le \frac{1}{\psi^*}\\
        \mu F_z \text{sgn}(\alpha),           & |\sigma_{y}| > \frac{1}{\psi^*}
    \end{cases} \\
    M_z &= 
    \begin{cases}
        \label{eq:Mz}
        -\mu F_z\ell\psi^*\sigma_{y} [1-3|\psi^*\sigma_{y}|+3(\psi^* \sigma_{y})^2-|\psi^*\sigma_{y}|^3], & |\sigma_{y}|\le \frac{1}{\psi^*}\\
        0,           & |\sigma_{y}| > \frac{1}{\psi^*}
    \end{cases}
\end{align}
}

The values $\frac{1}{\psi}$ and $\frac{1}{\psi^*}$ represent the boundary between the adhesion region ($\sigma\le \frac{1}{\psi}$), where the bristles deform without sliding, and the sliding region ($\sigma> \frac{1}{\psi}$), where the tire has reached its maximum available friction. Here, sgn is the sign function, $\sigma = \sqrt{\sigma_x^2+\sigma_y^2}$ and $\psi$ and $\psi^*$ are composite tire parameters:
\begin{equation}
\label{eq:theta_star}
    \psi=\frac{2c_{p}\ell^2}{3\mu F_z} \qquad
    \psi^* = \frac{\psi}{1 - \ell\varphi\psi\text{sgn}(\alpha)}
\end{equation}

The normal load on each tire is calculated through load transfer based on the tire's distance from the center of gravity and total mass of the MMV. The control inputs $\Omega$ and $\delta$ are also derived based on the tire's position. Lastly, $r$ and $\alpha$ are calculated from the MMV's current velocity and transformed into the tire coordinate frame. To calculate the total force on the MMV, the forces and moments of each tire are transformed into the vehicle body coordinate system and summed.



\section{Differentiable Dynamics of KBM and GM3}
Both KBM and GM3 consist of equations that are almost fully differentiable. By implementing differentiable formulations of these dynamics models, DiffKBM and DiffGM3, we can compute 
$\partial \mathcal{L}_j / \partial \rho_i$, the gradient of the training objective $\mathcal{L}_j$ 
of downstream module $j$ with respect to the parameters $\rho_i$ of upstream
module $i$. Differentiability enables us to learn all parameters jointly for the overall objective with gradient-based optimization. We can also view DiffKBM and DiffGM3 as a layer in a neural network that automatically fine-tunes these learnable parameters for MMV dynamics.

For example, we can define the loss function as the average displacement between the predicted trajectory $\hat{Y} = \{\hat{y}_1, \hat{y}_2, ..., \hat{y}_T \}$ and the ground truth trajectory $Y=\{y_1, y_2, ..., y_T\}$.
\begin{equation}
    \mathcal{L}_{\text{ADE}} = \frac{1}{T} \sum_{t=1}^{T} \left|\left|{\hat{y}_t - y_t}\right|\right|_2
\end{equation}
The predicted trajectory comes from the dynamics model $y_t=f(x_t, u_t, \rho)$ (either DiffKBM or DiffGM3) where $x_t$ is the state vector, $u_t$ is the control input, and $\rho$ is the set of physical parameters. Then, for each parameter, $\rho_i$, we update its value using gradient descent with learning rate $\eta$. 
\begin{equation}
    \rho_i := \rho_i - \eta \frac{\partial \mathcal{L}_{\text{ADE}}}{\partial \rho_i}
\end{equation}

We get $\frac{\partial \mathcal{L}_{\text{ADE}}}{\partial \rho_i}$ by using the chain rule.
\begin{equation}
   \frac{\partial \mathcal{L}_{\text{ADE}}}{\partial \rho_i} =  \frac{\partial \mathcal{L}_{\text{ADE}}}{\partial f(x_t, u_t, \rho)} \frac{\partial f(x_t, u_t, \rho)}{\partial \rho_i}
\end{equation}
\subsection{DiffKBM: Differentiable KBM}
To study differentiable dynamics for micro-mobility systems, we begin by taking the classical kinematic bicycle model (KBM)~\cite{polack2017kinematic} and reformulate it as a fully differentiable dynamics model that we call DiffKBM. We define the vehicle state at time $t$ as
\[
    x_t = [x_{t,x}, x_{t,y}, \theta_t, v_t]
\]
where $x_{t,x}$ and $x_{t,y}$ denote the planar position, $\theta_t$ is the heading angle, and $v_t$ is the velocity. The control input is
\[
    u_t = [a_t, \delta_t]
\]
where $a_t$ is longitudinal acceleration and $\delta_t$ is the steering angle. The continuous kinematic bicycle equations are
\[
    \dot{x} = v \cos \theta, \ \ \dot{y} = v \sin \theta, \ \ \dot{\theta} = \frac{v}{L} \tan(\delta), \ \ \dot{v} = a,
\]
with wheelbase $L$. We discretize these equations using Euler integration with timestep $\Delta t$:
\begin{align*}
    v_{t+1} &= \text{ReLU}(0, v_t + a_t \Delta t) \\
    \theta_{t+1} &= \theta_t + \frac{v_{t+1}}{L} \tan (\delta_t) \Delta t \\
    x_{t+1, x} &= x_{t,x} + v_t \cos(\theta_t) \Delta t \\
    x_{t+1, y} &= x_{t,y} + v_t \sin(\theta_t) \Delta t
\end{align*}

We clamp $v$ to be non-negative as we assume forward-only motion and treat $L$ as a learnable parameter. Since all operations in the update equations are differentiable almost everywhere, gradients are able to backpropagate through control inputs and the model parameters.

\begin{figure}
\vspace{5pt}
    \centering
    \includegraphics[width=\linewidth]{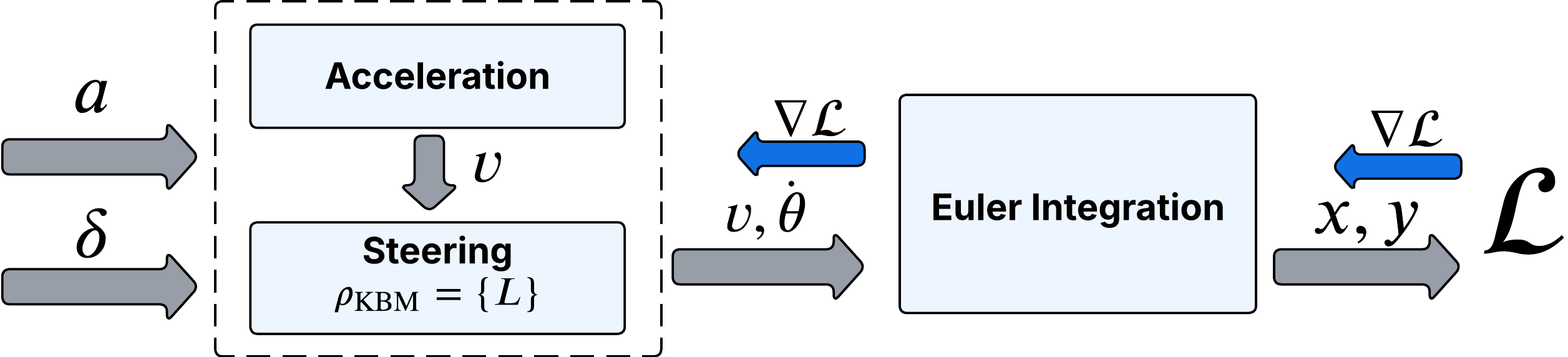}
    \caption{DiffKBM implements wheelbase $L$ as a learnable parameter in the steering component. The resulting turning rate $\dot{\theta}$ along with velocity are used in Euler integration to obtain the position. By making KBM end-to-end differentiable, we are able to tune $L$ for an MMV with unknown wheelbase length.}
    \label{fig:diffkbm}
\end{figure}

\subsection{DiffGM3: Differentiable GM3}
\newcommand{\norm}[1]{\left\lVert#1\right\rVert}

\begin{figure*}[th]
    \vspace{5pt}
    \includegraphics[width=\textwidth]{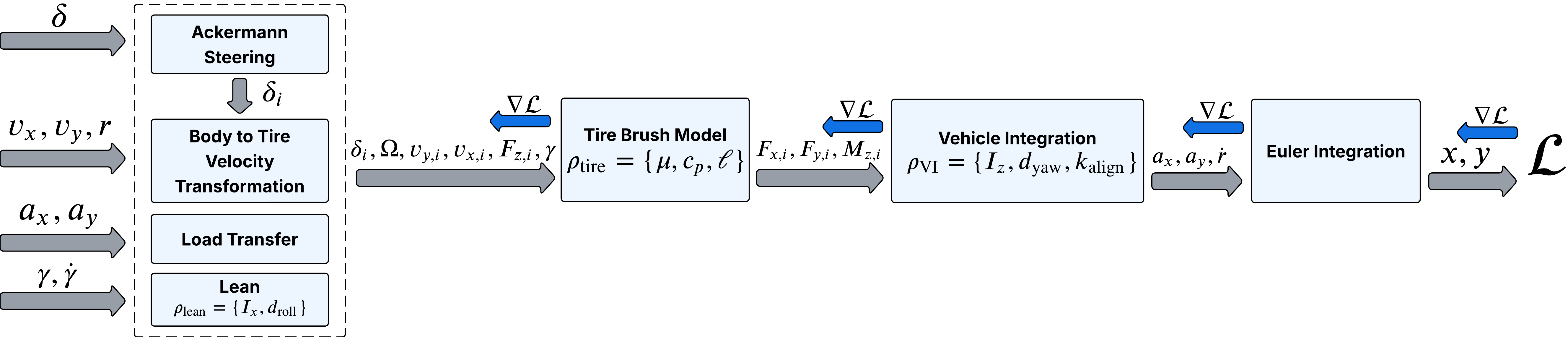}
    \caption{DiffGM3 contains several differentiable components with sets of learnable parameters $\rho_{\text{lean}}, \rho_{\text{tire}}$, and $\rho_{\text{VI}}$. First, the steering control input $\delta$ is converted into tire-specific steering angles $\delta_i$. This, along with the MMV's previous longitudinal velocity $v_x$, lateral velocity $v_y$, and yaw rate $r$, is used to calculate tire velocities. Meanwhile, longitudinal and lateral accelerations, $a_x$ and $a_y$, are used to transfer load among the tires, and the previous lean angle $\gamma$ and roll rate $\dot{\gamma}$ are used to determine the next lean angle. In addition to wheel angular velocity $\Omega$, the results from these modules are inputs to the tire brush model, where forces and moments are computed. The outputs for each tire are integrated into the vehicle dynamics to obtain longitudinal, lateral, and yaw acceleration. Finally, the resulting position is calculated with Euler integration. This fully differentiable formulation allows gradients to backpropagate to each module so parameters can be tuned on MMV trajectory data.}
    \label{fig:DiffGM3}
    \vspace*{-1em}
\end{figure*}

In this section, we present the detailed formulation on DiffGM3, including differentiable steering and lean for controlling a GM3 system, as shown in
Figure~\ref{fig:DiffGM3}.

DiffGM3 automatically ``discovers'' or learns several physical parameters that are not easily measurable: coefficient of friction, half contact length, cornering stiffness, and moments of inertia. We can tune these parameters on real-world MMV trajectory data by using a differentiable formulation of the equations used in GM3. 

The formulas for GM3 contain several discontinuities that make the current formulation non-differentiable at certain critical points. In particular, the formulas for longitudinal force, lateral force, and aligning moment introduced in \ref{sec:GM3} contain discontinuities at the boundary between the adhesion region and the sliding region when $\sigma = \frac{1}{\psi}$. Since the derivative of the sign function is 0 everywhere, we get vanishing gradients when differentiating Equation \ref{eq:Fy}. The absolute values in Equation \ref{eq:Mz} also cause non-differentiability when $\psi^*\sigma_y=0$. 

We can eliminate these discontinuities by replacing them with smooth functions. For discontinuities at boundaries, we implement a gate $G(\sigma, \psi) = \frac{1}{1 + e^{-k(\sigma - 1/\psi)}}$, where $k$ is the steepness, to transition between the adhesion region and sliding region. Then, we replace the sign function $\text{sgn}(\alpha)$ with $\tanh{(k\alpha)}$. For absolute values, we use $\hat{\sigma} = \sqrt{\sigma_x^2 + \sigma_y^2 + \epsilon}$, where $\epsilon << 1$. Finally, we add soft clamping using tanh to keep the parameters and control inputs within a physically realistic range. Letting $f(\sigma, \psi) = 3\psi \sigma - 3(\psi \sigma)^2 + (\psi \sigma)^3$ and $g(\sigma, \psi) = 1 - 3\psi \sigma + 3(\psi \sigma)^2 - (\psi \sigma)^3$, we have the following equations for DiffGM3:

{\footnotesize
\begin{align}
    F_x &= \left[ (1 - G(\hat{\sigma}, \psi))  \mu F_z \frac{\sigma_x}{\hat{\sigma}} f(\hat{\sigma}, \psi) \right] + \left[ G(\hat{\sigma}, \psi)  \mu F_z \frac{\sigma_x}{\hat{\sigma}} \right] \\
    F_y &= \left[ (1 - G(\hat{\sigma}, \psi^*))  \mu F_z \frac{\sigma_y}{\hat{\sigma}} f(\hat{\sigma}, \psi^*) \right] + \left[ G(\hat{\sigma}, \psi^*)  \mu F_z \tanh(k \alpha) \right] \\
    M_z &= (1 - G(\hat{\sigma}, \psi^*))  \left[ -\mu F_z \ell \psi^* \sigma_y  g(\hat{\sigma}, \psi^*) \right]
\end{align}
}%

Since there is a sign function in Equation \ref{eq:theta_star}, we replace it with $\psi^* = \frac{\psi}{1 - \ell \varphi\psi\tanh(k\alpha)}$.

\subsubsection{Differentiable Steering}
In the original Ackermann steering formula for a double-track vehicle, $\tan {\delta_i} = \frac{L}{R-y_i}$, where $\delta_i$ is the angle of tire $i$, $y_i$ is its lateral position, and $R=\frac{L}{\tan{\delta}}$ is the turning radius. This leads to a singularity at $\delta=0$. To make this differentiable, we use the following equation to calculate the steering angle $\delta_i$ of tire $i$ based on input steering angle $\delta$, where $|\delta| < \frac{\pi}{2}$.

\begin{equation}
    \delta_i = \arctan\left( \frac{L \tan(\delta)}{L - y_i \tan(\delta)}\right)
\end{equation}

Skateboards do not use Ackermann steering, and the equations for skateboard truck geometry are already fully differentiable.

To keep $\delta$ within a physically feasible range (handlebars on bicycles and scooters cannot rotate more than $\approx 45^\circ$ to the left or right), we use soft clamping with maximum steering angle $\delta_{\text{max}}$.


For the aligning moment, we introduce a damping factor $d_\text{yaw}$ to stabilize the turning dynamics. Thus, our yaw acceleration is
\begin{equation}
    \dot{r} = \frac{M_{z}^{\text{total}}}{I_z} - d_\text{yaw} r
\end{equation}
where $r$ is yaw rate and $I_z$ is moment about the z axis.
\subsubsection{Differentiable Lean}
To simulate differentiable leaning behavior, we implement a second-order dynamic model. Instead of calculating an equilibrium lean angle $\gamma = \arctan(v^2 / gR)$, DiffGM3 treats roll as a degrees-of-freedom influenced by mass and inertia.

The roll acceleration $\ddot{\gamma}$ is determined by the balance of torques around the longitudinal axis at the ground contact point:
\begin{equation}
I_x \ddot{\gamma} + d_{\text{roll}} \dot{\gamma} + m g h_{\text{cg}} \sin(\gamma) = m a_y h_{\text{cg}} \cos(\gamma)
\end{equation}
where $I_x$ is the roll moment of inertia, $a_y=v_x \cdot r$ is the lateral acceleration, and $d_{\text{roll}}$ is the damping coefficient for roll. To ensure the model remains stable during backpropagation, we use soft clamping with maximum lean angle $\gamma_{\text{max}}$.
This formulation ensures that $\frac{\partial \gamma_{out}}{\partial \gamma_{in}}$ never becomes zero, allowing the optimizer to recover if the vehicle ``falls over" during a training epoch. The resulting $\gamma$ after Euler integration is used to calculate turn slip in Equation \ref{eq:turn_slip}.

Our smoothing follows the same principle as differentiable relaxations in deep learning, where hard switches are replaced with sigmoid or tanh surrogates so gradients can flow, and the gate steepness $k$ acts as a temperature. Unlike a neural activation, it approximates a known physical model with a negligible gap. The brush model approximations only introduce an error near the adhesion–sliding transition, and the force curve is already twice continuously differentiable there. The resulting worst-case force error decays with $(1/k)^3$. The tanh replacement for the sign function and the softened norm is only inaccurate near zero slip, precisely where the force terms they affect are vanishing. The smoothed model is therefore the exact model plus a small bounded disturbance, and since receding-horizon MPC re-plans from the measured state at every step, this mismatch cannot accumulate.

\section{Implementation and Experiments}
We evaluate DiffGM3 in two settings: (1) open-loop trajectory matching on real-world MMV data using DiffStack~\cite{karkus2023diffstack}, and (2) closed-loop autonomous navigation in pedestrian crowds using differentiable MPC (DiffMPC)~\cite{amos2018differentiable}. The open-loop experiment evaluates how well DiffKBM and DiffGM3 can reproduce real MMV trajectories when used as the dynamics model inside a differentiable autonomy stack. The closed-loop experiment evaluates whether DiffGM3 enables safe and predictable navigation when used for online planning and control in pedestrian crowd scenarios. 

\textbf{Hardware.} For the open-loop experiments, we trained DiffStack for 50 epochs using 4 CPU cores, 8 NVIDIA RTX A4000 GPUs, and 4GB of memory, taking 1-2 hours.

\textbf{IMPTC Dataset.} 
We use the Infrastructural Multi-Person Trajectory and Context (IMPTC) dataset~\cite{hetzel2023imptc} to evaluate DiffGM3 in open-loop settings, which contains real-world trajectories of vehicles (cars and buses), MMVs (e.g. bicycles, scooters), and pedestrians collected from everyday public road traffic. The data was captured in different weather conditions and includes traffic light signal data. The full dataset includes 270 sequences, each containing $\approx 50$ vehicle tracks, totaling $\approx 13,500$ tracks.

\textbf{Parameter Optimization on IMPTC.} The control inputs required for GM3 are steering angle and wheel angular speed. To our knowledge, there is no existing MMV trajectory dataset with this information. As a workaround, we utilize DiffMPC to infer the control sequence that best reproduces each observed trajectory by solving a finite-horizon tracking problem. Because DiffGM3 is differentiable, we backpropagate trajectory tracking loss through time to optimize model parameters. Using 30 trajectories from the IMPTC dataset, we calibrate six dynamic parameters for realistic bicycle dynamics, as shown in Table~\ref{tab:paramOptim}.
\begin{table}[t]
\centering
\vspace{5pt}
\caption{DiffGM3 parameters optimized using IMPTC trajectories. Align gain $k_{\text{align}}$ and damping coefficients $d_{\text{yaw}}, d_{\text{roll}}$ help to stabilize the MMV and reduce oversteering. Moments of inertia $I_z$, $I_x$, and tire parameters are tuned because they are not easily measurable.}
\label{tab:paramOptim}
\begin{tabular}{ll}
\toprule
\textbf{Component} & \textbf{Optimized Parameters} \\
\midrule
Bicycle body & $k_{\text{align}}$, $d_{\text{yaw}}, d_{\text{roll}}$, $I_z$, $I_x$ \\
Tires (each wheel) & $\mu$, $c_p$, $\ell$ \\
\bottomrule
\end{tabular}
\vspace*{-2.5em}
\end{table}
Learning physical parameters from trajectory data is a form of system identification, and identifiability depends on the richness of the inputs. The IMPTC trajectories consist of low-speed intersection maneuvers, so the tires operate primarily in the adhesion regime: the effective cornering
stiffness is strongly excited, while friction-dominated parameters that govern behavior near saturation are only weakly excited, and the persistent
excitation condition is not fully satisfied for all parameters. To keep the optimization well-posed, we soft-clamp all physical parameters within
physically feasible ranges using $\tanh$, which acts as an implicit prior. 

\subsection{Open-loop Trajectory Matching Evaluation on IMPTC.}
To demonstrate GM3’s ability to be used in AV stacks for autonomous MMVs, we replace the dynamically-extended unicycle model in DiffStack with DiffKBM and DiffGM3. Since our new formulations are differentiable, they can directly replace the unicycle model and allow gradients to backpropagate through the control cost function and further to the prediction module. We use DiffKBM and DiffGM3 for the ego MMV only because we are focused on training the planner and controller to produce ideal plans. Within DiffStack, DiffMPC solves a finite-horizon optimal control problem at each planning step, rolling out candidate control sequences through the dynamics model (DiffKBM or DiffGM3) and minimizing the planning cost to produce the control inputs. Because both the dynamics and the cost are differentiable, the loss can be back-propagated through this optimization, allowing the control solution to inform the tuning of upstream physical parameters and cost weights.

\begin{figure*}[ht]
\centering
    \vspace{5pt}
        \begin{subfigure}[b]{0.2\textwidth}
                \includegraphics[width=\linewidth]{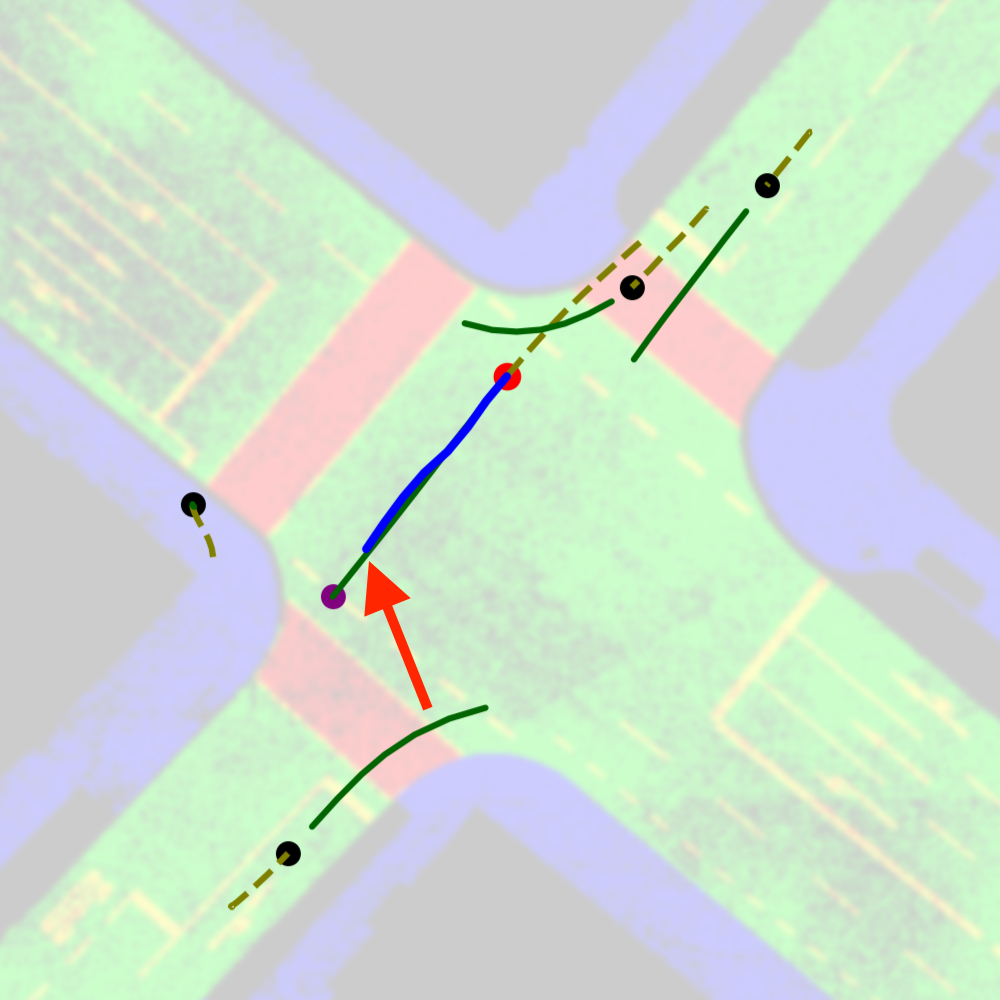}
                \caption{Bicycle: DiffGM3}
                \label{subfig:diffstack-diffgm3}
        \end{subfigure}%
        \begin{subfigure}[b]{0.2\textwidth}
                \includegraphics[width=\linewidth]{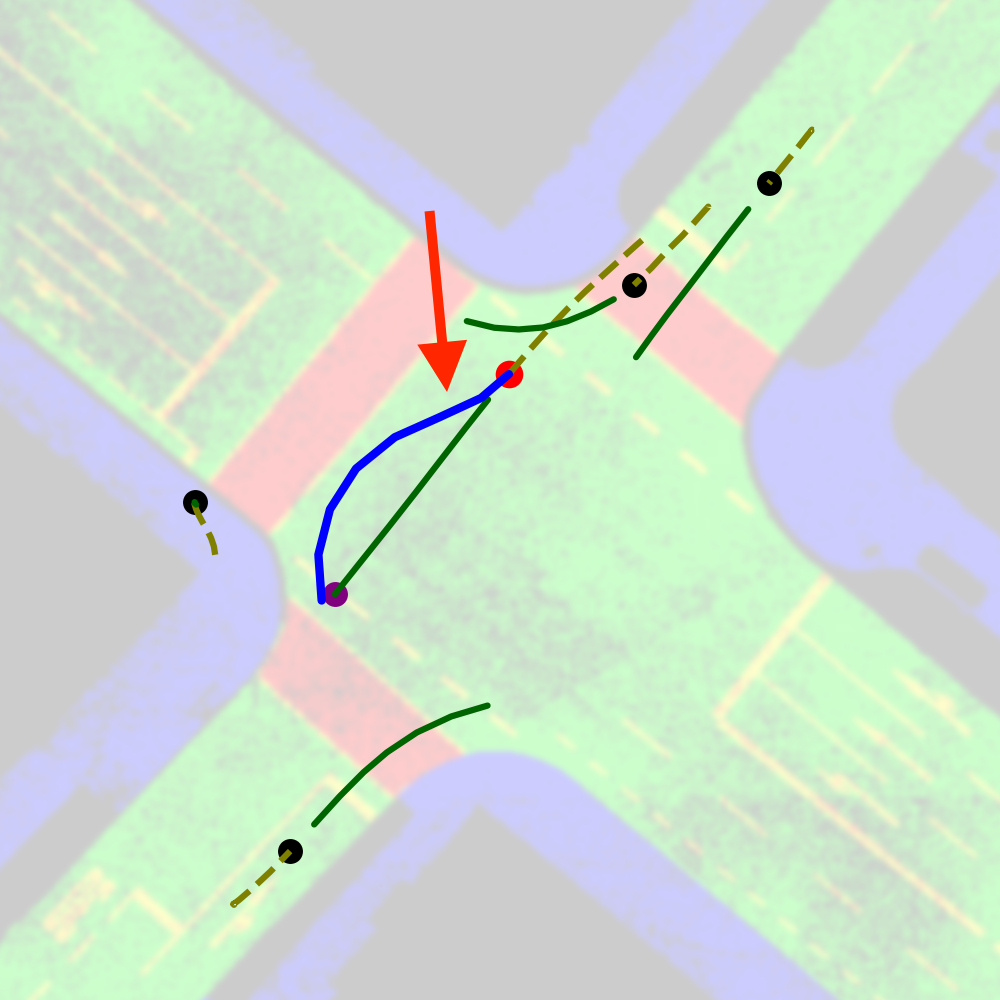}
                \caption{Bicycle: DiffKBM}
                \label{subfig:diffstack-unicycle}
        \end{subfigure}%
        \begin{subfigure}[b]{0.2\textwidth}
                \includegraphics[width=\linewidth]{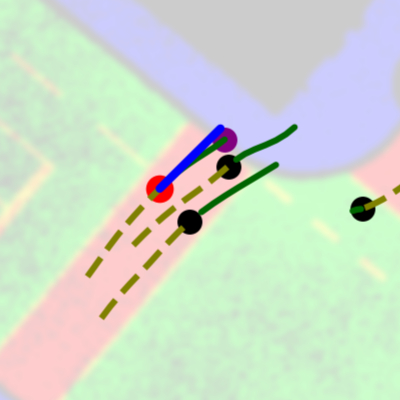}
                \caption{Scooter: DiffGM3}
                \label{subfig:diffstack-diffgm3}
        \end{subfigure}%
        \begin{subfigure}[b]{0.2\textwidth}
                \includegraphics[width=\linewidth]{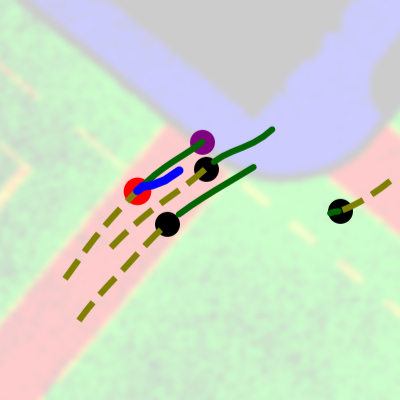}
                \caption{Scooter: DiffKBM}
                \label{subfig:diffstack-unicycle}
        \end{subfigure}%
        \begin{subfigure}[b]{0.15\textwidth}
                \includegraphics[width=\linewidth]{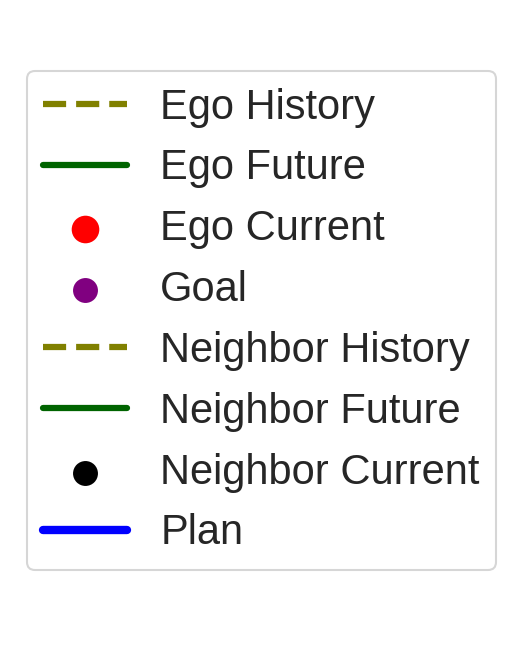}
                \label{subfig:diffstack-legend}
        \end{subfigure}%
        
        \vspace*{-0.5em}
        \caption{\textbf{Open-loop Trajectory Matching Comparison between the DiffKBM and DiffGM3}: (a) On bicycle trajectories, DiffGM3 is able to stay on path because it has learnable parameters to prevent overturning. However, in some cases, it stops before the goal as shown by the red arrow, leading to a higher goal cost in the planning loss function. (b) On the other hand, the DiffKBM reaches its goal but veers off the ground-truth straight path, as indicated by the red arrow. This is due to the DiffKBM's lack of support for realignment in its formula for yaw. (c) For a scooter, DiffGM3 is able to steer the scooter to its goal while avoiding other agents. (d) On the same trajectory, the DiffKBM causes the ego agent to turn towards another agent and slow down, preventing it from reaching its goal.}
        \label{fig:diffstack-results}
\end{figure*}

\textbf{Experimental Setting.} We conduct imitation learning (IL) experiments to reproduce MMV trajectories from the IMPTC dataset. The goal is to minimize the MSE between the output ego trajectory and the ground truth in the dataset. The cost function is a linear combination of the loss functions from each module $\mathcal{L} = \omega_1\mathcal{L}_{\text{pred}}+\omega_2\mathcal{L}_{\text{plan}}+\omega_3\mathcal{L}_{\text{ctr}}$. We experiment with different weight values. In $\mathcal{L}_{\text{plan}}$, we prioritize collision avoidance and goal cost. We configure Trajectron++ in DiffStack to predict the trajectories of at most 16 nearest agents within a 30-meter radius, including cars, pedestrians, and other MMVs. While IMPTC provides a segmentation map, it does not include vectorized semantic data, so we do not use any map data in the planner. 

For open-loop training, we sample planning scenarios from a subset of IMPTC with $H=4$s history and $T=3$s future data and select one MMV as the ego agent. We designate bicycles, scooters, and motorcycles as the ego MMV and all other agents for trajectory prediction. The full training set from the IMPTC dataset was used for training and includes 617 bicycle tracks, 233 motorcycle tracks, and 85 scooter tracks. We evaluate DiffStack on Average Displacement Error (ADE) and negative log-likelihood (NLL) for ego trajectories. We evaluate planning on cross-entropy loss. We compare the metrics from training DiffStack with DiffKBM and DiffGM3.

\textbf{Results.} The results of our open-loop experiments are summarized in Table \ref{tab:open}. DiffGM3 consistently outperforms DiffKBM across all modes on ADE and NLL. For ADE, it achieves reductions of 1.16\% for bicycles, 9.84\% for scooters, and 2.37\% for motorcycles. Similarly, NLL improves by 3.02\%, 2.05\%, and 4.75\%, respectively. For planning loss, DiffGM3 shows significant gains on bicycles (4.16\%) and motorcycles (28.9\%), though it underperforms on scooters with a 19.39\% increase in loss. This is due to DiffGM3 slowing down before the scooter reaches the goal. However, in these scenarios, DiffGM3 is able to stay on path while DiffKBM swerves off-path and has to turn back to reach the goal. Figure~\ref{fig:diffstack-results} shows a qualitative comparison between DiffKBM and DiffGM3. Since DiffGM3 has parameters to dampen yaw, it can correct for overturning. On the other hand, DiffKBM lacks parameters to minimize the accumulation of heading angle from yaw rate over time, resulting in deviation from the ground truth trajectory, even though it can reach its goal. This deviation is particularly pronounced in intersection environments, where MMVs are unlikely to make aggressive turns.

\begin{table*}[th]
    \centering
    \begin{tabular}{lcccccccc}
    \toprule
         & \multicolumn{2}{c}{$\downarrow$\textbf{ADE (m)}} & \phantom{a} &  \multicolumn{2}{c}{$\downarrow$ \textbf{NLL}} & \phantom{a} &  \multicolumn{2}{c}{$\downarrow$ \textbf{Planning Loss}} \\
        \cmidrule{2-3} \cmidrule{5-6} \cmidrule{8-9}
        \textbf{Mode} & \textbf{DiffGM3} & \textbf{DiffKBM} & \phantom{a} & \textbf{DiffGM3} & \textbf{DiffKBM} & \phantom{a} & \textbf{DiffGM3} & \textbf{DiffKBM}\\
        \midrule
        Bicycle  & $\mathbf{2.06 \pm 0.08}$ & $2.08 \pm 0.07$ & \phantom{a} & 
        $\mathbf{1.07 \pm 0.00}$& $1.10 \pm 0.04$  & \phantom{a} &
        $\mathbf{149.10 \pm 7.87}$ & $155.58 \pm 10.38$  \\
        Scooter & $\mathbf{1.45 \pm 0.01}$& $1.61 \pm 0.02$ & \phantom{a} & 
        $\mathbf{1.23\pm 0.19}$ & $1.28 \pm 0.17$ & \phantom{a}& $142.75 \pm 2.09$& $\mathbf{119.57 \pm 3.01}$  \\
        Motorcycle  & $\mathbf{1.74 \pm 0.00}$ & $1.78 \pm 0.06$ & \phantom{a} & $\mathbf{3.02 \pm 0.05}$ & $3.17 \pm 0.27$ & \phantom{a} & $\mathbf{212.23 \pm 12.77}$ & $298.50 \pm 7.54$  \\
        \bottomrule
    \end{tabular}
    \caption{\textbf{Open-loop Trajectory Matching Evaluation Results:} DiffGM3 outperforms DiffKBM in ADE (Bicycle: -1.16\%, Scooter -9.84\%, Motorcycle: -2.37\%) and NLL (Bicycle: -3.02\%, Scooter -2.05\%, Motorcycle: -4.75\%) on all modes. For planning loss, DiffGM3 outperforms DiffKBM on bicycle (-4.16\%) and motorcycle (-28.9\%) trajectories, but performs worse on scooters (19.39\%).}
    \label{tab:open}
    \vspace*{-2em}
\end{table*}

\subsection{Closed-loop Navigation in Pedestrian Crowds}
\begin{figure*}
    \centering
    \vspace{5pt}
    \includegraphics[width=0.95\textwidth]{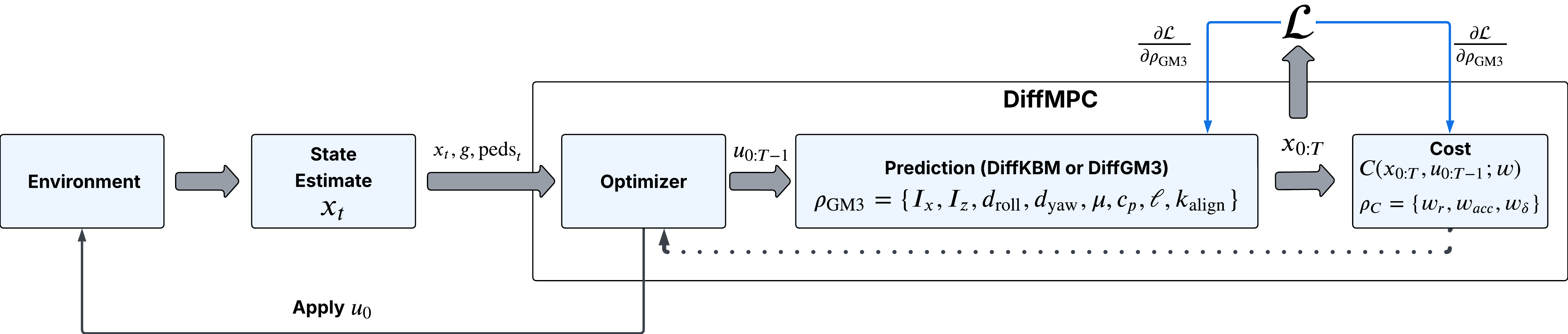}
    \caption{{\bf Overview of the DiffMPC pipeline integrated with DiffKBM or DiffGM3}. At each timestep, the environment provides a state estimate $x_t$, which is passed to the optimizer with the goal position and surrounding pedestrian states $\text{peds}_t$. The optimizer proposes a control sequence $u_{0:T-1}$ and uses either DiffKBM or DiffGM3 parameterized by learnable parameters $\rho_{\text{KBM}}$ or $\rho_{\text{GM3}}$ to roll out predicted future states $x_{0:T}$. A cost function is defined with learnable cost weights $\rho_C$. Because both the dynamics and the cost function are differentiable, gradients can be back-propagated through the prediction and cost modules to enable end-to-end learning of MMV dynamics parameters. The first control action $u_0$ is applied to the environment, and the process repeats in a receding-horizon loop.}
    \label{fig:DiffMPC+DiffGM3}
    \vspace*{-0.75em}
\end{figure*}

We evaluate closed-loop autonomous navigation using CrowdNav, a crowd-robot interaction simulation benchmark that provides circle-crossing and square-crossing scenarios in which an ego robot must reach a goal while avoiding collisions with multiple moving pedestrians \cite{chen2019crowd}. We pair DiffGM3 with a differentiable DiffMPC controller \cite{amos2018differentiable}. At each timestep, DiffMPC plans a finite horizon control sequence using DiffGM3 to predict future MMV motion and executes the first control action as seen in Figure~\ref{fig:DiffMPC+DiffGM3}. We similarly pair DiffKBM with a DiffMPC controller to serve as a baseline. For both systems, the planning horizon, cost terms, weights, and optimization settings were all chosen based on the recommendations given in the DiffMPC paper \cite{amos2018differentiable}; only the underlying dynamics model differs. We report success rate, collision rate, navigation time, discomfort frequency, and average minimum distance to pedestrians.

\textbf{Experimental Setting.}
The state of our robot is defined as $x = [p_x, p_y, v_x, v_y, \theta, r]$, representing positions, velocities, heading, and yaw rate, respectively. We control the robot using $u = [a, \delta]$, representing acceleration and steering angle. Our cost function $C$ emphasizes reaching the goal, avoiding collisions, and maintaining a safe distance from pedestrians. It is defined as
\begin{equation}
    C(x_{0:T},u_{0:T-1}) = \sum_{t=0}^{T-1} \left[C_{\text{goal}} (x_t) + C_{\text{cbf}}(x_t) + C_{\text{reg}}(x_t, u_t)\right]
\end{equation}
where the components are defined as:

\begin{enumerate}
    \item Goal tracking ($C_{\text{goal}}$): We force the robot towards the goal $g = [g_x, g_y]$ while keeping the heading towards it $\theta_{\text{des}}$ and avoiding overshooting the goal:
    {\footnotesize
    \begin{equation}
    C_{\text{goal}}(x_t) = w_{\text{pos}} ||p_t-g||^2 + w_\theta (\theta_t-\theta_{\text{des}})^2+w_{\text{vel}} ||v_t-v_{\text{ref}}(t)||^2
    \end{equation}
    }%
    We added a dynamic reference velocity $v_{\text{ref}}(t)$ using a physics-based braking profile to ensure the robot decelerates as it nears the goal.

    \item Collision avoidance ($C_{\text{cbf}}$): To account for the non-holonomic constraints of the bicycle model, we define a cost based on {\em Turning Circle Control Barrier Function} (TC-CBF) \cite{lee2025turningcirclebasedcontrolbarrier}: 
    \begin{equation}
        C_{\text{cbf}}(x_t) = w_{\text{coll}} \text{ReLU}(-h(x_t))^2
    \end{equation}
    where $h(x)$ defines if both turning left and right are obstructed.
     
    \item Regularization ($C_{\text{reg}}$): We add small penalties to ensure smooth control:
    \begin{equation}
        C_{\text{reg}}(x_t, u_t) = w_r r_t^2 + w_{\text{acc}} a^2_t + w_\delta \delta^2_t
    \end{equation}
\end{enumerate}

\begin{table*}[t]
\vspace{5pt}
\centering
\caption{{\bf Closed-loop Pedestrian Crowd Navigation Performance Comparison between DiffGM3 and DiffKBM:} DiffGM3 prioritizes safety and demonstrates significantly fewer collisions across all vehicle configurations relative to DiffKBM. Specifically, in the baseline comparison between the two bicycle models, DiffGM3 produces {\bf 55\%} fewer collisions, with a {\bf 3\%} better success rate. DiffGM3's incomplete rate and slower navigation time can be explained by its tendency to avoid collisions by veering away from the goal, resulting in situations where the vehicle is unable to arrive at the goal within the time limit. This also explains why DiffGM3's discomfort frequency is {\bf 50\%}-{\bf 75\%} better than DiffKBM's across all modes. Moreover, DiffKBM's faster navigation times are primarily the result of collisions ending the scenario early.}
\label{tab:kbm_gm3_results}
\begin{tabular}{lcccccc}
\toprule
\textbf{Model} & $\uparrow$ \textbf{Success Rate} &  $\downarrow$ \textbf{Collision Rate} & $\downarrow$ \textbf{Incomplete Rate} &  $\downarrow$ \textbf{Nav. Time (s)} & $\downarrow$ \textbf{Discomfort Freq.} & $\uparrow$ \textbf{Avg. Min Dist} \\
\midrule
DiffKBM Bicycle & 0.62 & 0.38 & \textbf{0.00} & \textbf{4.02} &  0.08 & \textbf{0.13} \\
DiffGM3 Bicycle & \textbf{0.64} & \textbf{0.17} & 0.19 & 5.13 & \textbf{0.02} & 0.12 \\
\hline
DiffGM3 Scooter & 0.63 & \textbf{0.13} & 0.23 & 6.14 & 0.03 & \textbf{0.13} \\
DiffGM3 Wheelchair & 0.60 & 0.23 & 0.17 & 6.30 & 0.03 & 0.12 \\
DiffGM3 Tricycle & \textbf{0.67} & 0.23 & 0.17 & 4.82 & 0.04 & 0.11 \\
\bottomrule
\end{tabular}
\vspace*{-1em}
\end{table*}

\begin{figure*}[h!btp]
    \vspace{5pt}
    \centering
    \begin{minipage}[c]{0.78\linewidth} 
        \centering
        \begin{subfigure}[b]{0.24\linewidth}
            \centering
            \includegraphics[width=\linewidth]{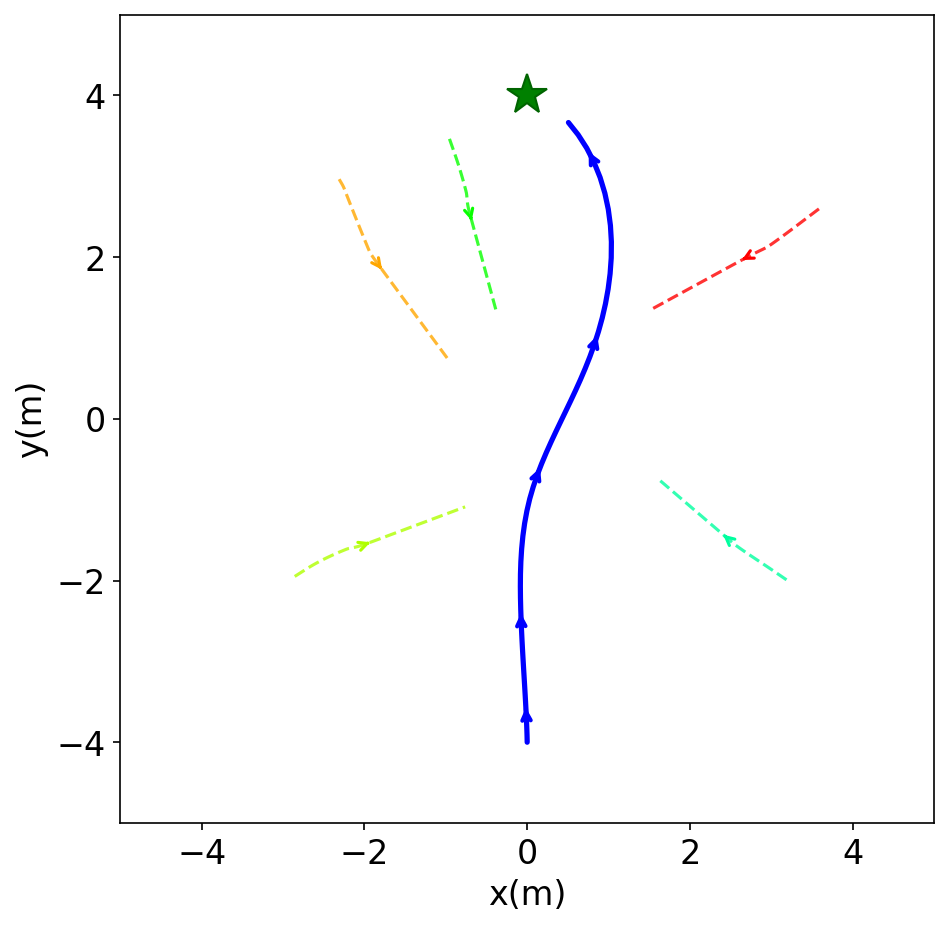}
            \caption{DiffGM3 Scenario 1}
        \end{subfigure}
        \begin{subfigure}[b]{0.24\linewidth}
            \centering
            \includegraphics[width=\linewidth]{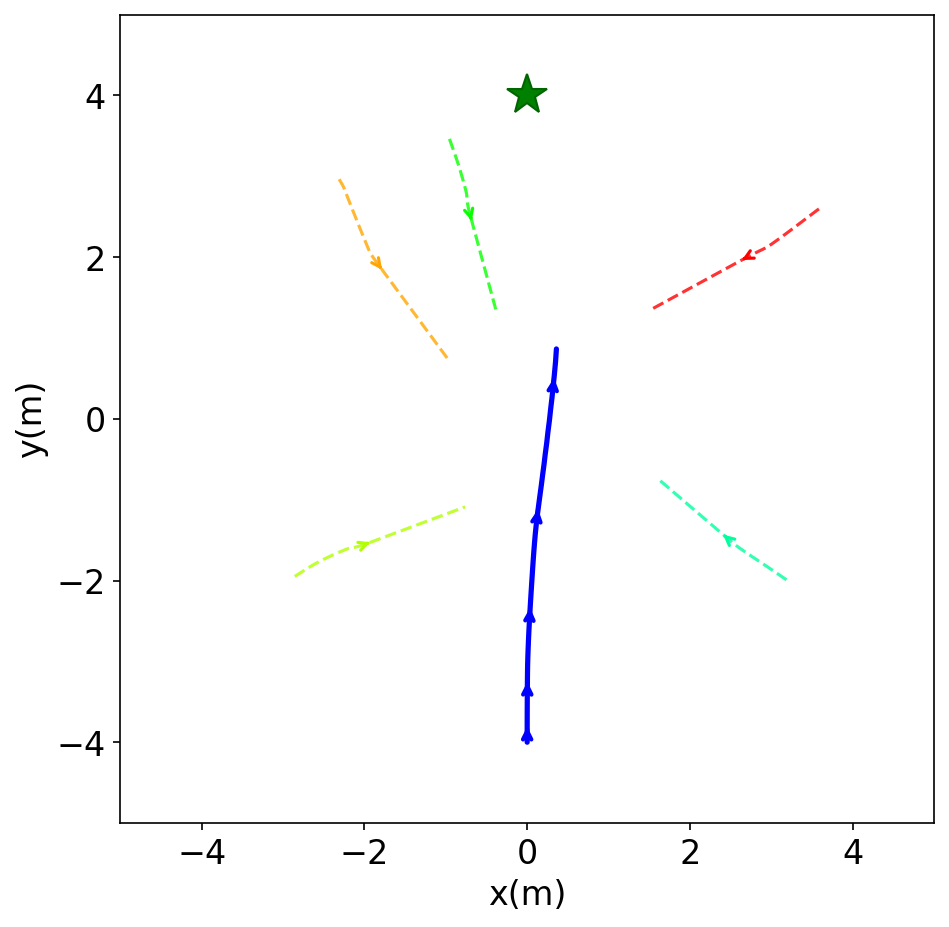}
            \caption{DiffKBM Scenario 1}
        \end{subfigure}
        \begin{subfigure}[b]{0.24\linewidth}
            \centering
            \includegraphics[width=\linewidth]{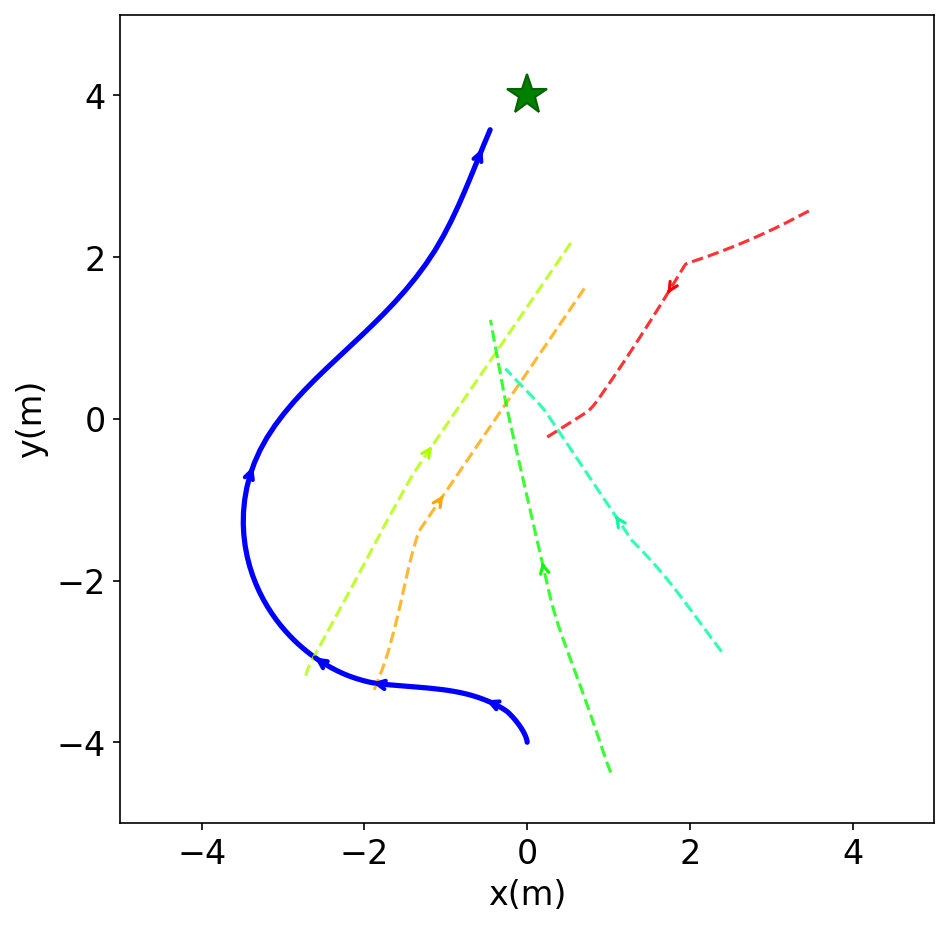}
            \caption{DiffGM3 Scenario 2}
        \end{subfigure}
        \begin{subfigure}[b]{0.24\linewidth}
            \centering
            \includegraphics[width=\linewidth]{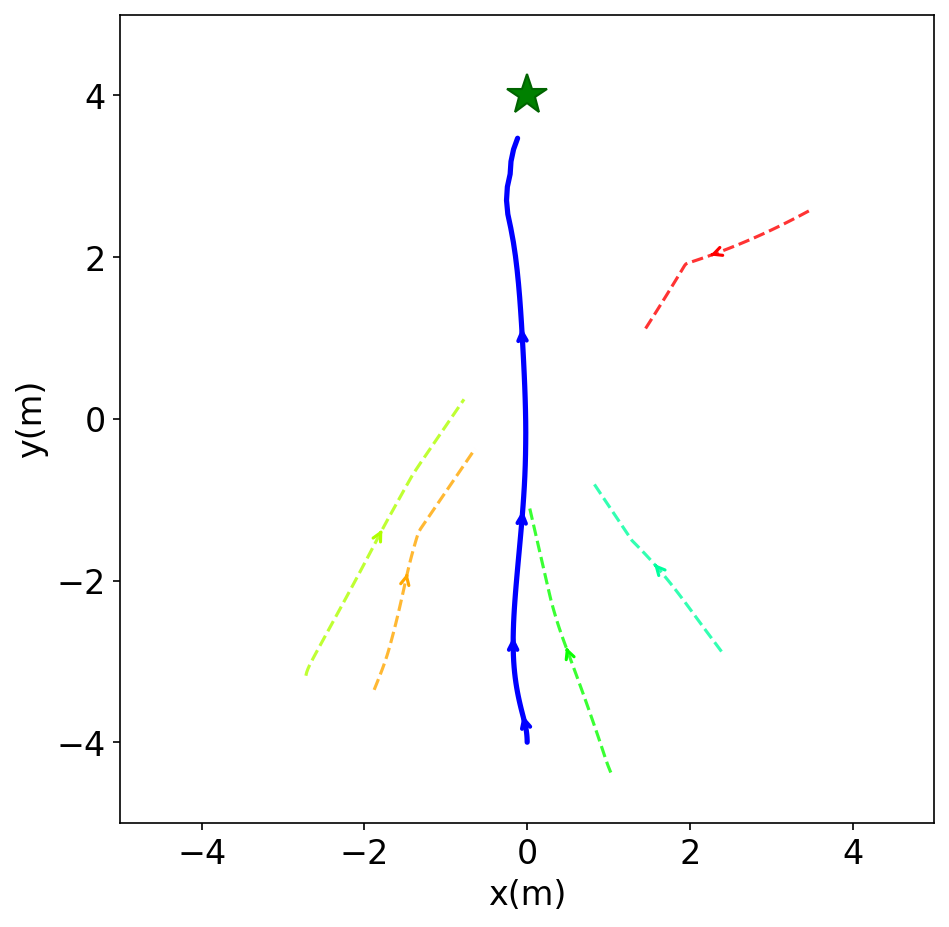}
            \caption{DiffKBM Scenario 2}
        \end{subfigure}
    \end{minipage}
    \hfill
    \begin{minipage}[c]{0.2\linewidth}
        \centering
        \includegraphics[width=\linewidth]{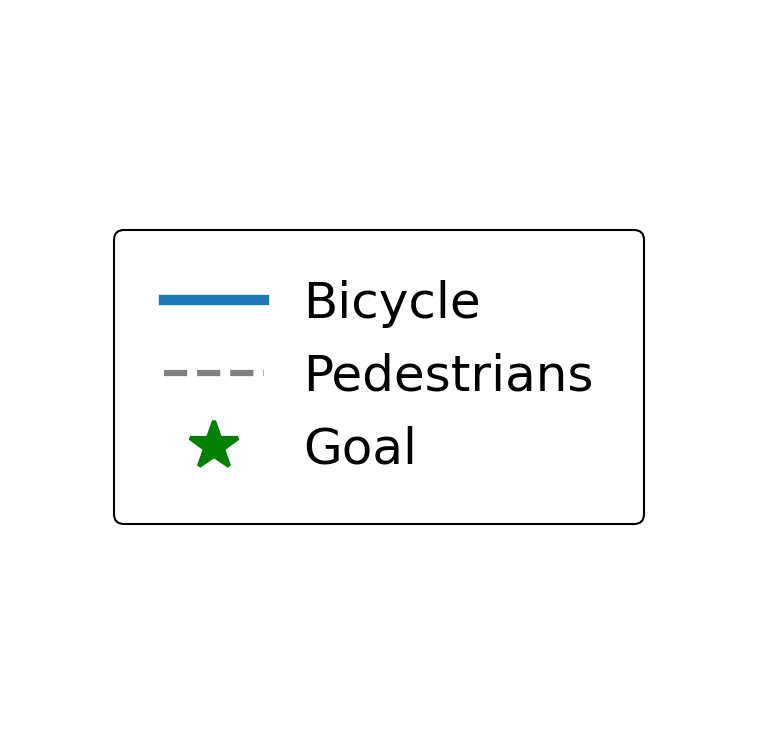}
    \end{minipage}

    \caption{\textbf{Closed-loop Navigation in Pedestrian Crowds Comparison between DiffGM3 Bicycle and DiffKBM:} We show two unique scenarios each for DiffGM3 and DiffKBM. Longer pedestrian paths reflect the longer navigation time taken by the MMV to avoid moving pedestrians to reach its goal.
    \textbf{Scenario 1:} DiffGM3's route reflects its ability to use real-world kinematics to avoid pedestrians when necessary, while DiffKBM's simple kinematics forces it to attempt to fit right in between two pedestrians, resulting in a collision.
    \textbf{Scenario 2:} While both models reach the goal, DiffGM3's longer route demonstrates {\em less discomfort and higher minimum distance} from pedestrians relative to DiffKBM's direct route.
    }
    \label{fig:gm3_kbm_closed_loop}
    \vspace*{-1em}
\end{figure*}

\textbf{Results.} We show the results after running both DiffKBM and DiffGM3 with multiple vehicle types on 100 sample trajectories from CrowdNav in Table~\ref{tab:kbm_gm3_results}. By keeping both systems constant besides the underlying dynamics model, we demonstrate the improvements DiffGM3 makes over the widely used DiffKBM. While DiffKBM's success rate is close to DiffGM3's, it has a significantly higher collision rate. The cases where DiffGM3 doesn't reach the goal include both collisions and incomplete routes within the time limit. This is due to DiffGM3's tendency to avoid pedestrians by turning away from the goal, causing it to be unable to turn back within the time limit of 10 seconds per run. We also show two common scenarios found during our comparison between DiffKBM and DiffGM3's bicycle model in Figure~\ref{fig:gm3_kbm_closed_loop}. As mentioned, DiffKBM's collision rate is significantly higher than all of our DiffGM3 MMVs. We attribute this to its simple kinematics model, which results in immediate acceleration and a very brute force approach to arrive at the goal. Specifically, we see many scenarios similar to scenario 1, where DiffKBM travels straight and finds itself being forced to squeeze in between two pedestrians. While it does succeed at times, it fails here and collides with them. DiffGM3, though, is able to utilize its real-world kinematics and apply braking and turning to make a path between the pedestrians. Scenario 2 involves both dynamics models reaching the goal, with DiffKBM using a faster and shorter route. However, routes like this are why its discomfort frequency is significantly worse than DiffGM3's. DiffGM3 is able to avoid all the pedestrians by taking a longer curved route around them, while DiffKBM cuts through the crowd.

\section{Conclusion} 
In this work, we explored differentiable formulations of micro-mobility dynamics by deriving DiffKBM and DiffGM3 from the kinematic bicycle model (KBM) and the General Micro-mobility Model (GM3). DiffGM3 provides a unified, tire-brush-based dynamics formulation for diverse micro-mobility vehicles, while enabling gradient flow through physically meaningful parameters. 

We evaluated DiffGM3 in both open-loop trajectory matching and closed-loop navigation in pedestrian-shared environments using DiffKBM as a baseline. We find that DiffGM3 improves the trajectories produced by DiffStack with DiffKBM by correcting for oversteering with damping coefficients. DiffGM3 also produces significantly safer and more direct trajectories towards a goal through a crowd of pedestrians compared to DiffKBM. Furthermore, the extensibility of DiffGM3 to other micro-mobility platforms such as scooters, tricycles, and wheelchairs suggests that unified differentiable MMV dynamics can serve as a strong foundation for end-to-end differentiable autonomy stacks. By explicitly modeling effects like tire slip, lean, and load transfer, DiffGM3 supports the development of autonomous MMV systems that are not only collision-free but also predictable and socially compliant in pedestrian-shared environments. As such, we believe DiffGM3 in particular provides a key building block toward scalable autonomy for diverse micro-mobility platforms.

\vspace*{0.25em}
\noindent
{\bf Limitations and Future Directions: } Our open-loop experiments were limited to a single intersection with short trajectories. We plan to evaluate on long-running trajectories in diverse situations, such as high-acceleration or high-curvature scenarios, which excite the saturation regime. We will also test DiffGM3 on other platforms (scooter, e-scooter, cart, wheelchair, etc), which would require trajectory data of these MMVs in diverse environments.
Furthermore, we plan to further extend this work to simulate complex mixed traffic and diverse interactions among pedestrians, vehicles, and independent MMV agents. Scenarios like these with intricate interactions will further demonstrate DiffGM3's ability to navigate safely and quickly to a goal destination. 
As another possible extension, we aim to integrate DiffGM3 into a differentiable simulator to allow gradients to flow through a combination of MMV, pedestrians, fleet autonomy, and highly complex interaction models, enabling joint optimization of all components in the environment.
Finally, we will deploy DiffGM3 to a physical platform, such as a four-wheeled robot, to demonstrate improved performance over DiffKBM in the real world.

\vspace*{0.1em}
\noindent
{\bf Acknowledgement: } This project is supported in part by Dr. Barry Mersky and Capital One E-Nnovate Endowed Professorships and UMD-ARL Cooperative Agreement.






\bibliographystyle{IEEEtran}
\bibliography{refs.bib}  

\end{document}